\documentclass[pmlr,twocolumn,10pt]{jmlr}
\mlhtrack{findings}
\newif\iffinal
\finaltrue
\iffinal
  \mlhanonymousfalse
  \ifmlhneedspmlr
    \jmlrvolume{XXX}\jmlryear{2026}
  \else
    \jmlrproceedings{}{ML4H 2026 - \mlhtrackname}
    \makeatletter
    \renewcommand*{\@titlefoot}{\scriptsize\copyright\ 2026
      \@jmlr@authors.\hfill\@reprint}
    \makeatother
  \fi
  \jmlrworkshop{Machine Learning for Health (ML4H) 2026}
\else
  \jmlrproceedings{}{Submitted to ML4H 2026: \mlhtrackname}
  \jmlrworkshop{Machine Learning for Health (ML4H) 2026}
\fi
\usepackage{booktabs}
\usepackage{siunitx}
\usepackage{flafter}
\usepackage{placeins}
\usepackage{float}
\iffinal\else
  \usepackage[switch]{lineno}
  \linenumbers
\fi
\renewcommand{\tableconts}[3]{%
  {\centering #3\par}%
  \vskip 3pt
  #2\label{#1}%
}

\title[HC-ROCP]{Certify, Report, or Refuse: Harm-Constrained Conformal Prediction for Clinical Actions}
\author{%
  \Name{Zhiyuan Li}\textsuperscript{1} \Email{zli2562@wisc.edu}%
  \AND
  \Name{Lequan Yu}\textsuperscript{2}\setcounter{footnote}{1}\thanks{Corresponding author.} \Email{lqyu@hku.hk}\\
  \addr \textsuperscript{1}College of Computing \& Artificial Intelligence, University of Wisconsin--Madison, Madison, WI, USA\\
  \textsuperscript{2}School of Computing and Data Science, The University of Hong Kong, Hong Kong SAR, China
}
\hypersetup{
  pdftitle={Certify, Report, or Refuse: Harm-Constrained Conformal Prediction for Clinical Actions},
  pdfauthor={Zhiyuan Li and Lequan Yu}
}
\begin{document}
\maketitle
\ifmlhdemo\else
\begin{abstract}
Routing a melanoma case to routine follow-up is a serious action error that diagnostic coverage alone does not rule out. We present Harm-Constrained Risk-Optimal Conformal Prediction (HC-ROCP), a training-free framework that uses frozen classifier probabilities and adds a harm penalty when choosing prediction sets and actions. We test the resulting policies on disjoint calibration splits with multiplicity adjustment. HC-ROCP certifies supported claims, reports descriptive fallback policies, and refuses unsupported subgroup claims. We evaluate these outputs retrospectively on COVID-19 radiographs, ISIC 2019 dermoscopy, and EyePACS-to-APTOS retinopathy transfer. Across ten calibration seeds, HC-ROCP reduces COVID-19 severe-action errors from the exact risk-optimal baseline's 4.68\% to 0.54\%. At a 5\% budget, global claims are certified for COVID-19 and ISIC, but the rare squamous-cell-carcinoma claim is refused. At 1\%, all settings receive descriptive reports. For each budget, we also report the minimum severe-case count needed to pass the global harm gate with zero errors, providing a starting point for further data collection.
\end{abstract}

\fi
\ifmlhneedsstatements
\paragraph*{Data and Code Availability}
We use publicly available medical-imaging benchmarks: COVID-19 chest radiographs \citep{chowdhury2020covid,rahman2021covqu}, ISIC 2019 skin-lesion images \citep{codella2019isic2018,tschandl2018ham10000,hernandezperez2024bcn20000}, and an EyePACS-to-APTOS diabetic-retinopathy transfer split \citep{kaggle2015eyepacs,kaggle2019aptos}. Instructions for obtaining and preprocessing the datasets, together with the complete code implementation, are available at \url{https://github.com/LZYEIL/HC-ROCP}.
\paragraph*{Institutional Review Board (IRB)}
This work does not require IRB approval as we analyze existing publicly available datasets.
\fi

\section{Introduction}\label{sec:intro}
Medical-image classifiers can inform decisions about routine follow-up, further testing, or specialist referral. A serious error occurs when a severe condition receives an inappropriate action, such as routing melanoma to routine follow-up. Conformal prediction provides sets of possible diagnoses with marginal label-coverage guarantees \citep{vovk2005algorithmic,shafer2008tutorial}. Risk-Optimal Conformal Prediction (ROCP) uses these sets to choose downstream actions under a robust loss criterion \citep{wang2026rocp}.

Neither label coverage nor average loss directly controls how often a severe case receives a harmful action. A set may contain the true diagnosis even when the selected action is harmful, and low average loss can conceal errors concentrated in a rare severe class. Low observed error also needs enough data to support a population-level claim. In our ISIC study, for example, zero errors among 42 squamous-cell-carcinoma cases are insufficient to certify an error rate below $5\%$ under our multiplicity correction. We therefore assess severe-action errors separately and test which global and subgroup claims the calibration data support.

Risk-controlling prediction sets, conformal risk control, and learn-then-test provide tools for evaluating prespecified risks \citep{bates2021rcps,angelopoulos2023crc,angelopoulos2021ltt}. We use these tools to connect two tasks: constructing policies that target severe-action errors and auditing the evidence for their coverage and harm claims. The audit supports three reporting decisions: \emph{certify} a supported claim, \emph{report} a descriptive fallback policy, or \emph{refuse} an unsupported subgroup claim.

We introduce Harm-Constrained Risk-Optimal Conformal Prediction (HC-ROCP), a framework based on frozen classifier probabilities. We extend ROCP's objective with a penalty for harmful actions, weighted by the predicted probability of a severe label. Varying this penalty and the coverage parameter produces a finite family of policies. Separate calibration splits test their coverage and harm rates with multiplicity adjustment. These tests determine the selected policy's global evidence label and which subgroup claims are supported. Figure~\ref{fig:example} shows an actual output. Section~\ref{sec:decision} explains these labels before introducing the optimization.

\begin{figure}[t]
\floatconts{fig:example}{\caption{An ISIC melanoma case and the output of HC-ROCP. This MEL case receives an urgent action, a descriptive policy label, and refusal of the MEL subgroup claim at $\delta=1\%$.}}{%
\begin{minipage}[c]{0.28\linewidth}
\includegraphics[width=\linewidth]{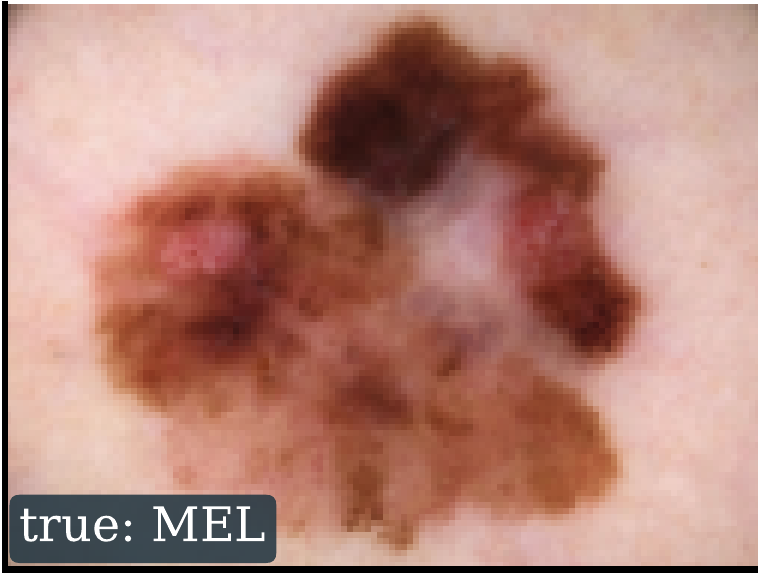}
\end{minipage}\hfill
\begin{minipage}[c]{0.68\linewidth}\small\raggedright
\textbf{Reference diagnosis:} MEL\\
\textbf{Set:} $\{\mathrm{MEL},\mathrm{SCC}\}$\\
\textbf{Modeled action:} urgent specialist pathway\\
\textbf{Policy:} expected-risk (descriptive)\\
\textbf{MEL subgroup:} insufficient evidence
\end{minipage}}
\end{figure}

We evaluate the method retrospectively in three imaging settings. On COVID-19, severe-action error is roughly ninefold lower than exact ROCP across ten calibration seeds, at higher modeled loss. ISIC tests whether aggregate evidence also supports a rare-subgroup claim. EyePACS-to-APTOS examines recalibration after a domain shift with few severe cases. No setting certifies at $1\%$. We vary the budget to identify which claims can pass and calculate the severe-case counts needed even with zero observed errors.

The paper makes three contributions:
\begin{itemize}
\item We extend ROCP with a penalty for prespecified severe-action errors, targeting them during the construction of prediction sets and actions (Section~\ref{sec:method}).
\item We jointly test coverage and harm when selecting a policy, report descriptive fallbacks, and withhold unsupported subgroup claims (Sections~\ref{sec:audit} and~\ref{sec:validity}).
\item We measure how subgroup size and candidate multiplicity affect certification, and how costs and referral options change the available policies (Section~\ref{sec:results}).
\end{itemize}

\begin{figure*}[!t]
\floatconts{fig:protocol}{\caption{Overview of the HC-ROCP framework. A construct split defines candidate policies; disjoint coverage and safety splits test their coverage and severe-action error before loss-based selection. The selected policy carries global and subgroup evidence labels.}}{\includegraphics[width=0.82\linewidth]{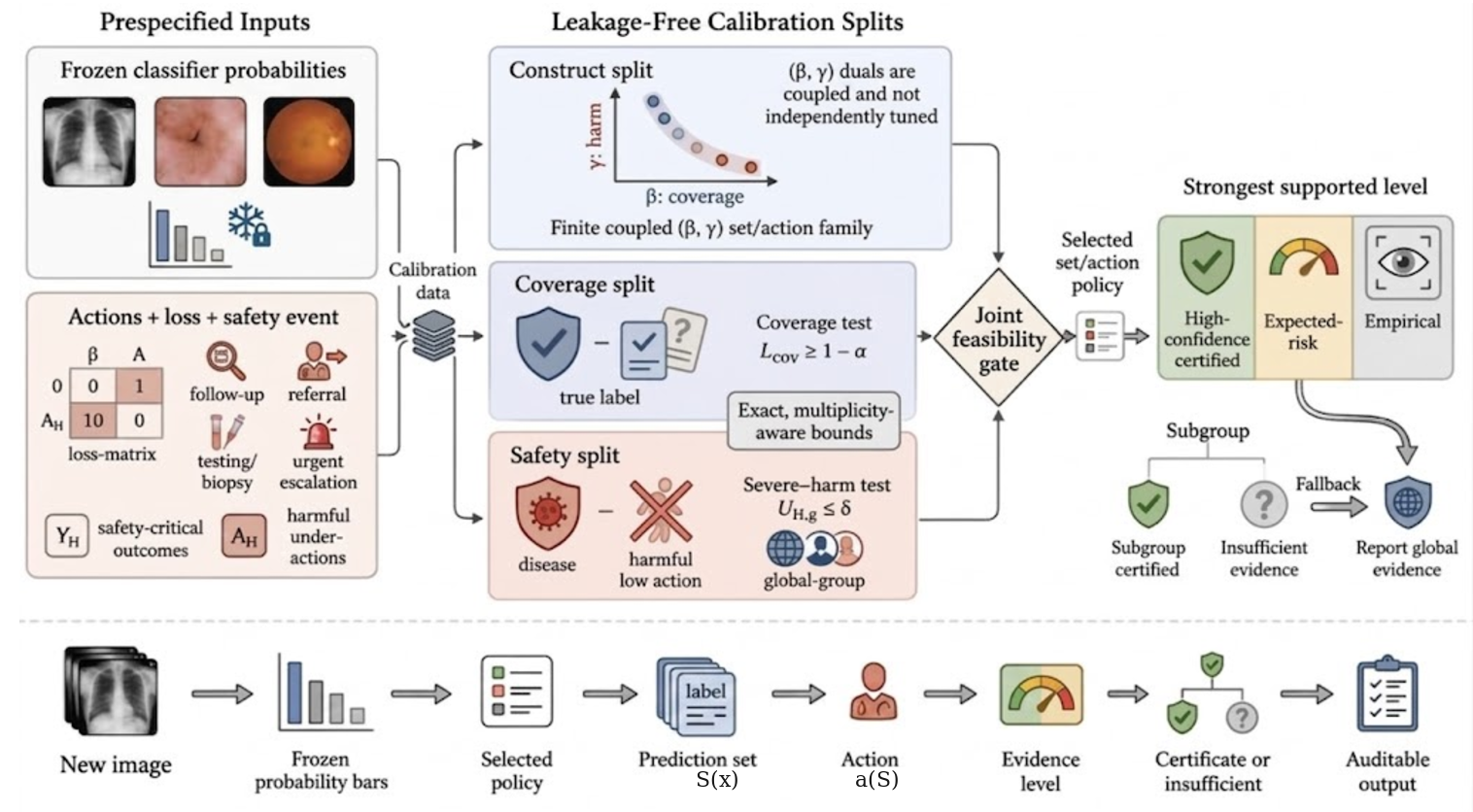}}
\vspace{-16pt}
\end{figure*}

\section{Related Work}
\paragraph{Coverage and Risk Control.}
Split conformal prediction controls marginal label coverage under exchangeability \citep{vovk2005algorithmic,shafer2008tutorial}. Risk-controlling prediction sets extend the target to losses \citep{bates2021rcps}, while conformal risk control and learn-then-test provide general tools for risk calibration and testing \citep{angelopoulos2023crc,angelopoulos2021ltt}. These tools can evaluate many kinds of decision policies. Our contribution integrates these bounds with harm-aware set/action construction and explicit global/subgroup reporting.

\paragraph{Decisions from Prediction Sets.}
ROCP optimizes a prediction set and its downstream action under a robust loss criterion \citep{wang2026rocp}. HC-ROCP preserves this action rule but adds a penalty for harmful actions on severe cases and tests their frequency alongside coverage. Conformal decision theory and utility-directed conformal prediction also connect uncertainty to decisions and utilities \citep{lekeufack2023cdt,cortesgomez2025udcp}. Conformal triage separates cases into high-risk, low-risk, and uncertain groups, with predictive-value guarantees for the first two \citep{angelopoulos2024triage}. Action-conditional conformal prediction defines risk conditional on the action taken \citep{zhu2026actionconditional}. We instead measure harmful-action frequency among cases whose true labels belong to the prespecified severe classes.

\paragraph{Joint Tests and Evidence Reporting.}
Related work studies joint certification across candidate policies \citep{yu2026jointcert} and hierarchical group-conditional control \citep{salem2026hgcrc}. \citet{gurram2026validityladder} distinguishes average-risk statements from high-confidence guarantees. \citet{liu2026certifyrefuse} study joint certification of error among accepted answers and a minimum answer rate under bounded-ratio covariate shift. Their procedure returns a refusal when no candidate passes both tests. HC-ROCP instead tests diagnostic label coverage and harmful-action frequency among severe cases over a finite, potentially non-nested policy family. We report global and subgroup evidence separately and retain descriptive fallbacks when joint certification fails.

\section{Problem and Outputs}\label{sec:decision}
\paragraph{The Decision Problem.}
An image $X$ has label $Y\in\{1,\ldots,K\}$ and frozen classifier probabilities $p(\cdot\mid X)$ (notation: Appendix~\ref{app:notation}). Before calibration, we specify actions $\mathcal A$, a loss matrix $\Lambda$, severe labels $Y_H$, and harmful actions $A_H$. A policy $\pi$ returns a nonempty set $S_\pi(X)$ and action $a_\pi(S_\pi(X))$. The severe-action-error event is
\[
 H=\mathbf 1\{Y\in Y_H,\ a_\pi(S_\pi(X))\in A_H\}.
\]
We require coverage $\mathrm{cov}(\pi)=P(Y\in S_\pi(X))\ge1-\alpha$ and conditional error
\[
 R_H(\pi)=P(a_\pi(S_\pi(X))\in A_H\mid Y\in Y_H)\le\delta.
\]
The denominator contains only severe cases. Modeled loss is $\mathcal L(\pi)=\mathbb E[\ell(a_\pi(S_\pi(X)),Y)]$, using $\Lambda$ to express relative action preferences.

\paragraph{Evidence Labels.}\label{sec:evidence}
A \emph{gate} tests a requested claim against the available calibration evidence (Table~\ref{tab:evidence}). HC-ROCP certifies a policy when coverage and all-severe harm gates pass together. Otherwise, it reports a fallback policy with its descriptive evidence level. Each subgroup claim receives its own harm test: passing supports certification, and failing produces a refusal. Disease-defined subgroup membership is determined from retrospective reference labels.

\begin{table}[t]
\floatconts{tab:evidence}{\caption{Evidence labels refer to policies or subgroups, not individual risk. Certification requires passing the applicable adjusted gates.}}{\small\setlength{\tabcolsep}{5pt}\begin{tabular}{p{0.32\linewidth}p{0.58\linewidth}}
\toprule
Label & Meaning\\
\midrule
High-confidence & Adjusted coverage and global harm gates pass.\\
Expected-risk & Add-one criteria pass; descriptive fallback.\\
Empirical & Raw-rate criteria pass; descriptive fallback.\\
Infeasible fallback & No preceding criteria pass; diagnostic only.\\
Subgroup: insufficient & Its adjusted harm gate fails; claim withheld.\\
\bottomrule
\end{tabular}}
\end{table}

\paragraph{A Running Example.}
For the melanoma image in Figure~\ref{fig:example}, the selected ISIC policy returns $\{\mathrm{MEL},\mathrm{SCC}\}$ and the \emph{urgent specialist pathway}. The policy carries an \emph{expected-risk} descriptive label, and the MEL subgroup's \emph{insufficient} label records refusal of its requested claim. The global label applies to the policy; the subgroup label applies to the MEL population. ISIC sets are larger on average (Section~\ref{sec:stability}).

\section{HC-ROCP: Construction and Audit}\label{sec:method}
We first construct a set/action family, then audit it on separate data (Figure~\ref{fig:protocol}; Algorithm~\ref{alg:protocol}).

\subsection{From a Prediction Set to an Action}
We retain ROCP's robust action rule. Let $\ell_{\rm in}(a;S)$ and $\ell_{\rm out}(a;S)$ be the largest losses inside and outside $S$, setting $\ell_{\rm out}=\ell_{\rm in}$ when the outside is empty. The robust risk is
\[
 \rho(S)=\min_{a\in\mathcal A}\!\left[\ell_{\rm in}(a;S)+\alpha_{\mathrm{act}}\{\ell_{\rm out}(a;S)-\ell_{\rm in}(a;S)\}_+\right].
\]
It bounds worst-case expected loss when at most $\alpha_{\mathrm{act}}$ mass lies outside $S$ \citep{wang2026rocp}. The minimizing action is $a(S)$, with fixed-order tie-breaking. Here $\alpha_{\mathrm{act}}$ defines the action rule, while $\alpha$ defines the audited coverage target. ROCP balances this risk against $p(S\mid x)=\sum_{y\in S}p(y\mid x)$:
\[
 S_\beta(x)\in\arg\min_{S\ne\emptyset}\{\rho(S)-\beta p(S\mid x)\}.
\]

\subsection{Constructing Harm-Aware Policies}
HC-ROCP selects a set minimizing
\[
 \rho(S)-\beta p(S\mid x)+\gamma p(Y_H\mid x)\mathbf 1\{a(S)\in A_H\}.
 \tag{1}
\]
The added term targets harmful actions on inputs assigned high severe-class probability. Its indicator activates only for sets whose robust action lies in $A_H$, while $p(Y_H\mid x)$ weights that penalty by the classifier's evidence of severity. The action map stays fixed and $\gamma=0$ recovers exact ROCP.

The two parameters act through the same set choice. Increasing $\beta$ rewards probability mass; increasing $\gamma$ favors sets whose associated action avoids $A_H$ on likely severe cases. A change in the preferred action can therefore change which labels enter the set. Jointly varying the parameters searches this coverage/action trade-off (Section~\ref{sec:joint_diagnostic}). We enumerate all nonempty subsets in the small label spaces studied here.

This objective generates candidates for
\[
\min_{\pi}\ \mathcal L(\pi)\quad\text{s.t.}\quad \mathrm{cov}(\pi)\ge1-\alpha,\ \ R_H(\pi)\le\delta .
\tag{2}
\]
Equation (1) is a pointwise, Lagrangian-inspired surrogate for (2); the audit in Section~\ref{sec:audit} tests the population constraints. Joint changes to the set and action produce a potentially non-nested family. We use a \emph{construct} split to fix the finite candidate family $\mathcal F$ before auditing.

\subsection{Joint Testing and Selection}\label{sec:audit}
For each candidate, we test whether coverage meets its target and severe-action error stays within budget. On a separate coverage split, $K_{\rm cov}(\pi)$ of $n_{\rm cov}$ labels fall inside the set. On the safety split, $K_{H,g}(\pi)$ of $n_{H,g}$ severe cases receive harmful actions in group $g$; group $0$ includes all severe labels.

For $M=|\mathcal F|$ policies and $G$ safety groups including group $0$, each test receives $\eta=\varepsilon/[M(1+G)]$. One-sided Clopper--Pearson tests produce coverage lower value $L_{\rm cov}$ and harm upper values $U_{H,g}$. A policy receives global certification if and only if
\[
 L_{\rm cov}(\pi)\ge1-\alpha
 \quad\text{and}\quad U_{H,0}(\pi)\le\delta.
\]
These gates are binomial-exact for a single class and conservative for stratified mixtures under Section~\ref{sec:validity}'s conditions. Among passing policies, we minimize safety-split loss. Simultaneous testing permits this selection on the safety data; utility itself receives no guarantee.

We use the construct split to propose parameter pairs or screen a shortlist, then freeze the family before inspecting coverage or safety outcomes. Because the simultaneous bounds cover every retained candidate, we can select any policy within that family. Adding candidates after inspecting failed gates would require a new validity argument. We use held-out test data to summarize the selected policy's performance. Only the calibration gates determine which certificate to issue.

\subsection{Descriptive Fallbacks and Subgroup Claims}
If no policy certifies, the expected-risk fallback requires
\[
 \frac{K_{\rm cov}+1}{n_{\rm cov}+1}\ge1-\alpha,
 \qquad \frac{K_{H,0}+1}{n_{H,0}+1}\le\delta.
\]
If none qualifies, the empirical fallback uses raw proportions. Both select the lowest-loss eligible policy. If neither is feasible, a diagnostic policy minimizes coverage shortfall, empirical harm, then loss. Empty severe groups cannot pass.

Both fallbacks are descriptive (Proposition~\ref{prop:ladder_main}). Each subgroup's adjusted harm gate determines its own certification status, regardless of the global label.

\section{What the Certificate Guarantees}\label{sec:validity}
A certificate controls false claims across the tested family for a fixed decision specification. Under the conditions below, with probability at least $1-\varepsilon$ over calibration, every certified harm claim meets its requested budget. The two quantities have distinct roles: $\varepsilon$ controls the probability of issuing a false claim, while $\delta$ bounds the action-error frequency among severe cases. Proofs are in Appendix~\ref{app:proofs}.

We condition on the construct split so that the candidate family is fixed, and on the class counts produced by stratified splitting. Observations are i.i.d.\ within each class, but error probabilities can differ across classes. Accordingly, $\mathrm{cov}_w$ and $R_{H,g,w}$ average class-conditional coverage and harm using the realized certification-split proportions (Appendix~\ref{app:proofs}).

For a single-class subgroup, the count is binomial. For a global mixture, we use the tail comparison below to justify gates computed from the binomial-reference Clopper--Pearson values $\mathrm{CP}_L$ and $\mathrm{CP}_U$. The $M(1+G)$ correction accounts for one coverage test and $G$ safety tests for each of the $M$ policies.

\begin{theorem}[Joint finite-sample certification]\label{thm:joint}
Let $\mathcal F$ have size $M$ and be fixed without inspecting certification outcomes, conditional on the realized class counts. For each policy and group, form the Clopper--Pearson gate values $L_{\rm cov}$ and $U_{H,g}$ with $\eta=\varepsilon/[M(1+G)]$. Assume every nonempty binomial-reference gate cutoff lies at least one count beyond its null mean, as formalized in Lemma~\ref{lem:hoeffding-tail}. Then, with probability at least $1-\varepsilon$, simultaneously over $\mathcal F$, passing $L_{\rm cov}(\pi)\ge1-\alpha$ implies $\mathrm{cov}_w(\pi)\ge1-\alpha$, and passing $U_{H,g}(\pi)\le\delta_g$ implies $R_{H,g,w}(\pi)\le\delta_g$ for every group $g$. Thus selection preserves the global statements and every separately passing subgroup statement.
\end{theorem}

\paragraph{Empirical Check.}
Gate-cutoff checks confirm the separation condition throughout (Appendix~\ref{app:proofs}); Appendix~\ref{app:protocolstudy} gives a simplified simulation of global harm tests.

\paragraph{Proof Sketch.}
Conditional on class counts, a global count is a sum of classwise binomials. Under the stated separation, Hoeffding's comparison bounds its gate tail by the binomial tail with the same mean \citep{hoeffding1956successes}; Clopper--Pearson inversion and a union bound over $M(1+G)$ gates complete the result.

\begin{proposition}[Add-one estimate]\label{prop:ladder_main}
Conditional on fixed stratified class counts, let $\pi$ be fixed without inspecting safety outcomes and write $K_H=\sum_{i=1}^{n_H}Z_i$, where the $Z_i$ are independent Bernoulli variables with possibly different probabilities $p_i$. For $p_w=n_H^{-1}\sum_i p_i=R_{H,0,w}(\pi)$,
\[
\mathbb E\!\left[\frac{K_H+1}{n_H+1}\right]=\frac{n_Hp_w+1}{n_H+1}\ge p_w.
\]
The analogous identity holds for coverage indicators and $\mathrm{cov}_w(\pi)$. Thus both add-one estimates are upward biased for a fixed policy, but neither is a conditional bound after observing its count or selecting $\pi$ from $\mathcal F$.
\end{proposition}

\paragraph{Proof Sketch.}
Since $\mathbb E[K_H]=\sum_i p_i=n_Hp_w$, the displayed identity follows by linearity of expectation; identical reasoning applies to coverage. The expectation is taken before observing the count for the fixed policy.


\begin{theorem}[Subgroup certify-or-fallback]\label{thm:subgroup}
Under the multiplicity correction of Theorem~\ref{thm:joint} applied to the global group and every requested subgroup, on the simultaneous-validity event: (i) every subgroup $g$ with $U_{H,g}(\pi^*)\le\delta_g$ is certified; (ii) a subgroup that fails the gate receives no subgroup certificate and the separately certified global policy remains valid (fallback); (iii) the realized risk of a failed subgroup is never claimed to be at most $\delta_g$.
\end{theorem}

\paragraph{Proof Sketch.}
Apply Theorem~\ref{thm:joint} with groups equal to the global group plus all requested subgroups. The simultaneous event includes the global statement and every passing subgroup statement, so withholding a failed subgroup claim preserves the certified global statement.

\paragraph{Choice of Bound.}
The binomial-reference bound is useful in the rare-event regime: for SCC's 42 cases with zero errors and $\eta\approx3.7\times10^{-5}$, its upper value is $21.6\%$, compared with $34.8\%$ for Hoeffding and $62.0\%$ for empirical-Bernstein at the same test level.

\section{Experimental Setup}\label{sec:setup}
\paragraph{Datasets and Decisions.}
We evaluate retrospective, single-stage policies against observed disease labels. COVID-19 radiographs \citep{chowdhury2020covid,rahman2021covqu} test the action-error/loss trade-off: the three abnormal labels are severe, and taking \emph{no action} on these cases is harmful. ISIC 2019 dermoscopy \citep{codella2019isic2018,tschandl2018ham10000,hernandezperez2024bcn20000} tests rare-subgroup evidence: MEL and SCC are severe, and \emph{routine follow-up} is harmful. EyePACS-to-APTOS retinopathy transfer \citep{kaggle2015eyepacs,kaggle2019aptos} tests target-domain recalibration: the two highest grades are severe, and \emph{routine rescreen} is harmful. Action sets and expert-specified loss matrices appear in Appendix~\ref{app:actions}; partitions are in Table~\ref{tab:sizes}.

\paragraph{Models and Calibration.}
All classifiers use ConvNeXt V2-Tiny \citep{woo2023convnextv2} and are frozen before HC-ROCP calibration, which requires no further classifier training. Class-stratified assignment produces disjoint construct, coverage, and safety splits. We set $\alpha_{\mathrm{act}}=\alpha=0.05$, family-wise error allowance $\varepsilon=0.05$, and reference budgets $\delta\in\{1\%,2\%,5\%\}$. Five clinicians contributed to the decision specifications and endorsed these budgets as reference points (Appendix~\ref{app:actions}).

Candidate $\beta$ values target coverage $0.95$, $0.975$, and $0.99$, crossed with the $\gamma$ grid. The higher targets provide room for finite-sample uncertainty, so joint selection can yield coverage above the required $0.95$.

\paragraph{Interpreting the Costs.}
The loss matrices encode relative preferences among available actions. For MEL, costs of $10$, $4$, $1$, and $0$ rank routine follow-up, dermatology referral, biopsy workup, and the urgent specialist pathway. We define the severe-action-error event separately from these costs, so we can vary the costs without changing which errors we count.

\paragraph{Comparisons and Metrics.}
Exact ROCP uses the same subset enumeration with $\gamma=0$; the original release solver uses a different threshold-based implementation \citep{wang2026rocp}. We also compare RAC, APS, RAPS, SAPS, Mondrian APS, LAS, SOCOP, utility-directed CP, Bayes action, always-safe action, and CRC/LTT action adapters \citep{kiyani2025rac,romano2020aps,angelopoulos2021raps,huang2024saps,vovk2013conditional,sadinle2019las,wang2025socop,cortesgomez2025udcp}. Appendix~\ref{app:implementation} documents official-code and formula adapters; action-only methods have no set coverage.

All methods share frozen probabilities, losses, and test observations. Tables~\ref{tab:matched} and~\ref{tab:baselines} compare full protocols (calibration details in Appendix~\ref{app:baselines}). Table~\ref{tab:baselines} summarizes ten-seed baseline comparisons; Table~\ref{tab:ablation} compares protocol variants on the reference split. We report coverage, mean modeled loss, severe-action-error rate, and evidence labels. Ten-seed means and sample standard deviations quantify calibration-partition variation on fixed test sets. Table~\ref{tab:main}, budget sweeps, and subgroup counts use one reference split. Paired tests are detailed in Appendix~\ref{app:baselines}; 500 nested-bootstrap repetitions are described in Appendix~\ref{app:ablation}.

\section{Results}\label{sec:results}
Three findings organize the results: HC-ROCP reduces severe-action error at a modeled-loss cost; global and subgroup audits support different claims; and budget, severe-case counts, and candidate multiplicity determine the evidence required for certification.

\begin{table}[!t]
\floatconts{tab:main}{\caption{Single-split selected operating points at $\delta=1\%$, with descriptive evidence labels. Seed-averaged comparisons appear in Tables~\ref{tab:matched} and~\ref{tab:baselines}.}}{\small\setlength{\tabcolsep}{3pt}\begin{tabular}{@{}lrrr@{}}
\toprule
 & COVID-19 & ISIC 2019 & \shortstack{EyePACS\\$\rightarrow$APTOS}\\
\midrule
Coverage & 0.9898 & 0.9477 & 0.9638\\
Harm & 0.50\% & 0.87\% & 0.00\%\\
Loss & 1.080 & 1.488 & 1.623\\
Evidence & expected-risk & expected-risk & empirical\\
$\beta$ & 35.72 & 11.93 & 28.88\\
$\gamma$ & 1.778 & 1.778 & 0.000\\
\bottomrule
\end{tabular}}
\end{table}

\begin{table}[t]
\floatconts{tab:matched}{\caption{Severe-action-error rate (mean $\pm$ sample standard deviation over ten calibration seeds). $\dagger$ denotes a Holm-significant reduction of at least 1 percentage point (Appendix~\ref{app:baselines}).}}{\resizebox{\linewidth}{!}{\begin{tabular}{lcc}
\toprule
Data & Exact ROCP ($\gamma{=}0$) & HC-ROCP\\
\midrule
COVID-19 & $4.68\%\pm0.45\%^{\dagger}$ & $0.54\%\pm0.32\%$\\
ISIC 2019 & $1.84\%\pm0.27\%^{\dagger}$ & $0.66\%\pm0.22\%$\\
EyePACS$\rightarrow$APTOS & 0.00\% $\pm$ 0.00\% & 0.00\% $\pm$ 0.00\%\\
\bottomrule
\end{tabular}}}
\end{table}

\subsection{COVID-19: Lower Error at Higher Modeled Loss}
Across ten seeds, HC-ROCP reduces severe-action error from exact ROCP's $4.68\%$ to $0.54\%$, while loss increases from $0.391$ to $1.181$. This approximately $8.7$-fold error reduction costs less than always taking the safe action, whose loss is $2.935$.

The reference policy has test harm $0.50\%$ and loss $1.080$ (Figure~\ref{fig:covid}). Its adjusted safety upper bound is $4.69\%$ from 366 severe cases, yielding an expected-risk report at $1\%$. The official CRC action adapter reaches $0.14\%$ harm at loss $1.534$ on this split; it reports an action-only output (Figure~\ref{fig:covid}).

\begin{figure}[t]
\floatconts{fig:covid}{\caption{Lower severe-action error has a modeled-loss cost. The single-split COVID-19 frontier shows candidate operating points and baseline policies. Shading marks test harm below $1\%$; the selected policy's calibration upper bound is $4.69\%$, so its evidence at this budget is descriptive.}}{\includegraphics[width=\linewidth]{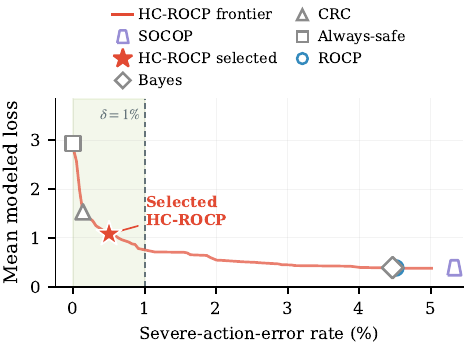}}
\end{figure}

\subsection{ISIC: Aggregate Evidence Leaves SCC Unsupported}\label{sec:isic}
On ISIC, the reference policy has test harm $0.87\%$ and an expected-risk label at the $1\%$ budget. The subgroup bounds are much wider. MEL has 302 safety cases and an upper bound of $4.88\%$, compared with $21.56\%$ for SCC's 42 cases. Neither subgroup passes at $1\%$. At $5\%$, the global and MEL claims pass, but the SCC claim is still refused (Figure~\ref{fig:isic}; all subgroups in Figure~\ref{fig:matrix}).

Ten-seed mean harm falls from exact ROCP's $1.84\%$ to $0.66\%$. The release solver has numerically lower harm and loss at different coverage; Appendix~\ref{app:baselines} reports the paired comparisons. The $5\%$ certified ISIC policy selects $\gamma=0$, with test coverage $0.9766$, harm $0.47\%$, and loss $1.631$.

\begin{figure}[t]
\floatconts{fig:isic}{\caption{MEL certifies at $5\%$, while SCC remains refused at both budgets. Bounds use the reference ISIC split and correspond to the policy selected at each budget.}}{\includegraphics[width=\linewidth]{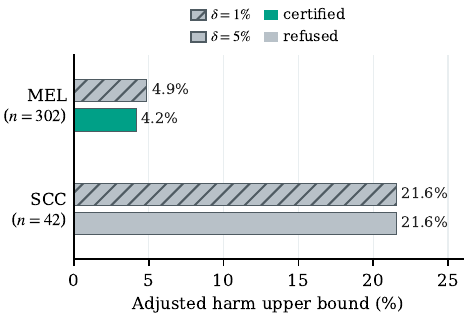}}
\end{figure}

\begin{samepage}
\subsection{EyePACS to APTOS: Zero Errors, Limited Evidence}
On EyePACS-to-APTOS, all methods have zero test harm. HC-ROCP selects $\gamma=0$, with ten-seed mean loss $1.705$ versus exact ROCP's $1.514$. The joint search includes $\gamma=0$, allowing it to retain an unpenalized policy. We also observe zero safety errors for the selected reference policy, but this is based on only 65 severe cases. At its adjusted test level $\eta$, the upper bound is $1-\eta^{1/65}=13.22\%$, well above the $1\%$ budget. The policy receives an empirical fallback label (Table~\ref{tab:main}).\par
\end{samepage}

\subsection{Budgets and Evidence Requirements}\label{sec:power}
For COVID-19, we select the same policy at the $1\%$ and $5\%$ budgets. Its calibration bound passes only the latter. In this sweep, global claims are certified for COVID-19 and ISIC only at $5\%$. EyePACS-to-APTOS returns descriptive reports at all three budgets (Figure~\ref{fig:power}).

The zero-error case gives an optimistic sample requirement. For zero errors among $n_H$ severe cases, the binomial-reference upper value is $1-\eta^{1/n_H}$. The minimum count for this gate to meet a budget $\delta$ is
\[
 n_0=\left\lceil\frac{\log\eta}{\log(1-\delta)}\right\rceil.
\]
Table~\ref{tab:sample_requirements} evaluates this threshold under the reference full-family Bonferroni correction. At $1\%$, all three settings have fewer severe cases than this zero-error threshold. Nonzero errors and the separate coverage and subgroup gates impose further requirements.

\begin{table}[t]
\floatconts{tab:sample_requirements}{\caption{Minimum severe-case counts for the zero-error global harm gate under the reference full-family correction. Coverage and subgroup gates impose further requirements.}}{\small\setlength{\tabcolsep}{3pt}\begin{tabular}{lrrrr}
\toprule
Data & Available $n_H$ & $1\%$ & $2\%$ & $5\%$\\
\midrule
COVID-19 & 366 & 1233 & 614 & 242\\
ISIC 2019 & 344 & 1015 & 505 & 199\\
EyePACS$\to$APTOS & 65 & 918 & 457 & 180\\
\bottomrule
\end{tabular}}
\end{table}

A construct-only shortlist lowers COVID-19's zero-error requirement from 1,233 cases for the full family to 459 for one candidate with the same coverage and subgroup claims. Both exceed the available 366 cases (Table~\ref{tab:shortlist}). Aggressive screening can also discard a policy that performs better on the disjoint safety split.

Figure~\ref{fig:power}(a) combines synthetic single-policy power curves with a separate post-selection count diagnostic on real COVID-19 severe cases. The real-data series yields zero gate passes; its sampling procedure is described in Appendix~\ref{app:ablation}.

\begin{figure*}[t]
\floatconts{fig:power}{\caption{Certification depends on the budget and available evidence. (a) Lines: synthetic single-policy power at $\delta=1\%$ over 5,000 repetitions. Crosses: post-selection count diagnostics from COVID-19 severe-case subsampling. Both use an unadjusted one-sided $95\%$ Clopper--Pearson harm gate. (b) Global evidence status under the full protocol.}}{\includegraphics[width=0.85\linewidth]{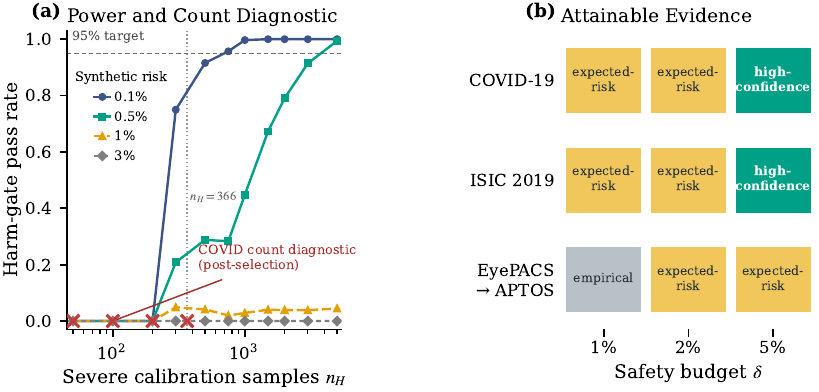}}
\end{figure*}

The official LTT action adapter likewise certifies zero policies at $1\%$, then certifies 16 COVID-19 and 7 ISIC candidates at $5\%$ (Table~\ref{tab:ltt}).

\subsection{Set Size, Costs, and Referral}\label{sec:stability}
\paragraph{Set Size.}
Set size counts the diagnosis labels returned for an image. We use these sets primarily to choose actions. Mean sizes across ten seeds are $1.732\pm0.185$ of four labels for COVID-19, $6.172\pm0.404$ of eight for ISIC, and $4.351\pm0.139$ of five for retinopathy. Each exceeds exact ROCP's mean (Table~\ref{tab:baselines}), quantifying the set-size cost alongside coverage.

\paragraph{Cost Scale.}
We halve the costs assigned to harmful actions on severe COVID-19 cases and repeat the full procedure. No empirically feasible candidate remains in any of the ten seeds, and the diagnostic fallbacks have harm $3.05\%\pm0.41\%$ (Figure~\ref{fig:loss}). With the original and doubled costs, harm is $0.54\%\pm0.32\%$ and $0.55\%\pm0.32\%$, respectively. Both retain expected-risk labels.

\paragraph{Referral.}
We add a referral action, modeled as non-harmful, and vary its cost. On the reference split, modeled loss falls from $1.080$ to $0.28$--$0.40$. At the lowest cost, however, the policy refers every case (Appendix~\ref{app:refer}). We report referral rates alongside loss to distinguish selective referral from referring every case.

\subsection{Why Joint Selection Matters}\label{sec:joint_diagnostic}
The fixed-$\beta$ diagnostic loses coverage: calibrating $\beta$ on the construct split at $\gamma=0$ and then choosing $\gamma$ by a safety-count rule gives $78.6\%$ test coverage, versus $99.0\%$ under joint selection (Appendix~\ref{app:ablation}).

Nested bootstrap retains each dataset's original global evidence label in all 500 repetitions while the selected parameters vary.

\section{Discussion and Limitations}
\paragraph{Certify, Report, and Refuse.}
Across the three datasets, we obtain global certificates for COVID-19 and ISIC at $5\%$ and descriptive reports for all settings at $1\%$. The SCC subgroup claim remains unsupported even when ISIC's global claim passes. We report observed performance for all selected policies, including those whose requested claims fail the certification tests.

Testing more candidate policies increases the multiplicity correction and raises the zero-error sample requirement. Costs and available actions instead change which policies are constructed. Future work will compare frozen classifiers under matched decision specifications and calibration budgets, improve construct-only screening, and develop reassessment procedures for workflow changes. The same certification tests can be applied to other prespecified policy families with appropriate selection control.

\paragraph{Path to Clinical Evaluation.}
Local clinicians would specify actions, harmful events, costs, and acceptable budgets, followed by auditing on representative labeled data. A prospective silent-mode study could assess action errors, subgroup evidence, and projected referral volume before recommendations enter care. An approved clinician-supervised study would subsequently evaluate overrides, escalation delays, workload, and patient outcomes. Hospital-specific review would also determine whether available referral capacity supports the modeled policies.

\paragraph{Limitations.}
First, we evaluate modeled actions against retrospective labels. The results therefore say nothing about actual patient outcomes or clinician workload. Expert-specified costs have not been validated as clinical utilities. The single-stage models omit treatment sequences, capacity constraints, and overrides, and COVID-19 actions may co-occur.

Second, calibration seeds share classifiers and test observations. Validity requires within-class i.i.d.\ sampling and the gate-cutoff conditions of Section~\ref{sec:validity}, and targets the realized calibration mixtures. Changes in prevalence or workflow require reassessment.

Third, the objective has no cardinality penalty, so large prediction sets can limit diagnostic specificity. Exact enumeration scales exponentially with label count, and robustness across classifier backbones remains to be evaluated.

\section{Conclusion}
We presented HC-ROCP, which constructs prediction sets and actions to target severe-action errors and tests the evidence for global and subgroup claims. Across three imaging settings, global claims are certified for COVID-19 and ISIC at $5\%$, but SCC remains unsupported and all settings return descriptive reports at $1\%$. Under the full-family correction, even zero observed errors would require more severe cases than are available at $1\%$, giving practitioners a concrete starting point for further data collection.

\clearpage
\bibliography{ref_new}
\clearpage
\onecolumn
\raggedbottom
\appendix
\section*{Appendix Contents}
\begingroup
\setlength{\parindent}{0pt}
\newcommand{\appendixentry}[2]{%
  \noindent\hyperref[#1]{\ref*{#1}\quad #2}\nobreak\dotfill\nobreak\pageref{#1}\par\medskip
}
\appendixentry{app:notation}{Notation}
\appendixentry{app:proofs}{Full Proofs}
\appendixentry{app:implementation}{Implementation Details}
\appendixentry{app:sensitivity}{Loss-Matrix Sensitivity}
\appendixentry{app:baselines}{Complete Baseline Comparison and Statistical Tests}
\appendixentry{app:ablation}{Protocol Diagnostics and Parameter Sensitivity}
\appendixentry{app:protocolstudy}{Simplified False-Certification Simulation}
\appendixentry{app:refer}{Referral Actions and the Action-Space Gap}
\endgroup

\clearpage
\section{Notation}\label{app:notation}

\begin{table}[H]
\floatconts{tab:notation}{\caption{Notation.}}{\begin{tabular}{ll}
\toprule
Symbol & Meaning\\
\midrule
$X$, $Y\in\{1,\ldots,K\}$ & image and its label\\
$p(\cdot\mid X)$ & frozen classifier probability vector\\
$S(X)$, $a(S)$ & prediction set and selected action\\
$Y_H$, $A_H$ & severe outcomes and harmful actions\\
$H=\mathbf{1}\{Y\in Y_H,\ a(S(X))\in A_H\}$ & severe-action-error event\\
$\delta$, $1-\alpha$ & harm budget and coverage target\\
$\pi=(S_\pi,a_\pi)$ & set/action policy\\
$\mathrm{cov}(\pi)$, $R_H(\pi)$ & population coverage and severe-action-error rate\\
$\mathrm{cov}_w(\pi)$, $R_{H,g,w}(\pi)$ & design-weighted certification targets\\
$\Lambda$ & modeled decision-loss matrix\\
$\rho(S)$ & robust set risk\\
$\alpha_{\mathrm{act}}$ & worst-case mass fraction outside $S$ in the action rule\\
$\beta$, $\gamma$ & coverage and harm parameters\\
$\varepsilon$ & family-wise error allowance\\
$M$ & candidate family size\\
$G$ & safety-group count, including the global group\\
$\eta$ & per-test error $\varepsilon/[M(1+G)]$\\
$\mathrm{CP}_L$, $\mathrm{CP}_U$ & binomial-reference Clopper--Pearson values\\
\bottomrule
\end{tabular}}
\end{table}

\section{Full Proofs}\label{app:proofs}
\paragraph{Clopper--Pearson Bounds.} For $K\sim\mathrm{Binomial}(n,p)$, the exact one-sided lower bound
\[
\mathrm{CP}_L(k,n,\eta)=\inf\{p:\Pr_{p}(\mathrm{Bin}(n,p)\ge k)\ge\eta\}
\]
equals the $\eta$-quantile of the Beta distribution $\mathrm{Beta}(k,n-k+1)$ (for $k\ge1$, and $0$ otherwise), and satisfies $\Pr(p\ge\mathrm{CP}_L)\ge1-\eta$. The exact upper bound
\[
\mathrm{CP}_U(k,n,\eta)=\sup\{p:\Pr_{p}(\mathrm{Bin}(n,p)\le k)\ge\eta\}
\]
equals the $(1-\eta)$-quantile of $\mathrm{Beta}(k+1,n-k)$ (for $k<n$, and $1$ otherwise), and satisfies $\Pr(p\le\mathrm{CP}_U)\ge1-\eta$.

\paragraph{Stratified Targets and Gate Cutoffs.} Let $n_{{\rm cov},y}$ be the fixed number of coverage-split observations with label $y$, and let $n_{H,g,y}$ be the fixed number of safety-split observations from severe class $y$ in group $g$. The conditional targets are
\[
\mathrm{cov}_w(\pi)=\sum_y\frac{n_{{\rm cov},y}}{n_{\rm cov}}\Pr(Y\in S_\pi(X)\mid Y=y),
\]
\[
R_{H,g,w}(\pi)=\sum_{y\in\mathcal Y_{H,g}}\frac{n_{H,g,y}}{n_{H,g}}\Pr(a_\pi(S_\pi(X))\in A_H\mid Y=y).
\]
For a binomial-reference coverage target $q$ and harm target $r$, define
\[
k_\eta(n,q)=\min\{k:\mathrm{CP}_L(k,n,\eta)\ge q\},\qquad
c_\eta(n,r)=\max\{k:\mathrm{CP}_U(k,n,\eta)\le r\}.
\]
If either defining set is empty, the corresponding gate cannot pass.

\begin{lemma}[Poisson--binomial tail comparison]\label{lem:hoeffding-tail}
Let $Z_1,\ldots,Z_n$ be independent Bernoulli variables with means $p_1,\ldots,p_n$, let $K=\sum_iZ_i$, and let $\bar p=n^{-1}\sum_i p_i$. For an integer $k\ge n\bar p+1$,
\[
 \Pr(K\ge k)\le\Pr\{\mathrm{Binomial}(n,\bar p)\ge k\};
\]
for an integer $c\le n\bar p-1$,
\[
 \Pr(K\le c)\le\Pr\{\mathrm{Binomial}(n,\bar p)\le c\}.
\]
These are the separated-tail comparisons of \citet{hoeffding1956successes}; the one-count conditions are essential. For their use below, we require a nonempty coverage gate at null value $q$ to satisfy $k_\eta(n,q)\ge nq+1$, and a nonempty harm gate at null value $r$ to satisfy $c_\eta(n,r)\le nr-1$.
\end{lemma}

\paragraph{Proof of Theorem~\ref{thm:joint}.} Fix $\pi$ and condition on the construct split and the stratified class counts. Write $C_{y,j}(\pi)$ for the coverage indicator of the $j$th coverage observation in class $y$. These variables are independent, with a common mean within each class, and
\[
K_{\rm cov}(\pi)=\sum_y\sum_{j=1}^{n_{{\rm cov},y}}C_{y,j}(\pi),\qquad
\mathbb E[K_{\rm cov}(\pi)]=n_{\rm cov}\mathrm{cov}_w(\pi).
\]
Passing the coverage gate is equivalent to $K_{\rm cov}(\pi)\ge k_\eta(n_{\rm cov},q)$. If $\mathrm{cov}_w(\pi)<q$ and the gate can pass, the assumed cutoff satisfies $k_\eta\ge n_{\rm cov}q+1>n_{\rm cov}\mathrm{cov}_w(\pi)+1$. Lemma~\ref{lem:hoeffding-tail}, monotonicity of a binomial upper tail in its success probability, and the definition of $\mathrm{CP}_L$ give
\[
\Pr\{L_{\rm cov}(\pi)\ge q\}
\le\Pr\{\mathrm{Bin}(n_{\rm cov},\mathrm{cov}_w(\pi))\ge k_\eta\}
\le\Pr\{\mathrm{Bin}(n_{\rm cov},q)\ge k_\eta\}
\le\eta.
\]
Similarly, within group $g$, $K_{H,g}(\pi)$ is a sum of classwise binomials with mean $n_{H,g}R_{H,g,w}(\pi)$. Passing its gate is equivalent to $K_{H,g}(\pi)\le c_\eta(n_{H,g},\delta_g)$. If $R_{H,g,w}(\pi)>\delta_g$, the second cutoff condition, Lemma~\ref{lem:hoeffding-tail}, lower-tail monotonicity, and $\mathrm{CP}_U$ inversion yield
\[
\Pr\{U_{H,g}(\pi)\le\delta_g\}
\le\Pr\{\mathrm{Bin}(n_{H,g},R_{H,g,w}(\pi))\le c_\eta\}
\le\Pr\{\mathrm{Bin}(n_{H,g},\delta_g)\le c_\eta\}
\le\eta.
\]
For a single-class subgroup, the corresponding Poisson--binomial distribution is binomial and the comparison is equality. Direct evaluation of the integer cutoffs verifies the separation conditions for every setting evaluated under Theorem~\ref{thm:joint}; the largest per-test error is $0.01$. Each of the $M(1+G)$ false-passing events has probability at most $\eta=\varepsilon/[M(1+G)]$; a union bound gives
\[
\Pr\Big(\exists\pi:\text{ a violated criterion passes its gate}\Big)\le M(1+G)\eta=\varepsilon.
\]
Hence, with probability at least $1-\varepsilon$, every passing coverage or group-safety gate supports its corresponding design-weighted criterion. Because this event is simultaneous over the candidate family, selecting $\pi^*$ by certification-split utility leaves the global and separately passing subgroup statements intact; utility itself receives no guarantee.

\paragraph{Proof of Proposition~\ref{prop:ladder_main}.} Conditional on the stratified counts, fix $\pi$ without inspecting the safety outcomes. Independence and linearity give $\mathbb E[K_H]=\sum_i p_i=n_Hp_w$. Therefore
\[
\mathbb E\!\left[\frac{K_H+1}{n_H+1}\right]
=\frac{n_Hp_w+1}{n_H+1}
=p_w+\frac{1-p_w}{n_H+1}\ge p_w.
\]
Replacing the harm indicators by coverage indicators gives the same identity with $p_w=\mathrm{cov}_w(\pi)$.

\paragraph{Proof of Theorem~\ref{thm:subgroup}.} Run the construction of Theorem~\ref{thm:joint} with the safety groups equal to $\{\text{global}\}\cup\{\text{requested subgroups}\}$, so $G=1+|\text{subgroups}|$ and the per-test error is $\varepsilon/[M(1+G)]$. On the simultaneous gate-validity event, $U_{H,g}(\pi^*)\le\delta_g$ implies $R_{H,g,w}(\pi^*)\le\delta_g$. If a subgroup fails the gate, the algorithm withholds its certificate and reports the global status; the global design-weighted coverage and safety statements hold on the same event regardless of the subgroup decisions. No statement about $R_{H,g,w}(\pi^*)$ is made for a refused subgroup.

\section{Implementation Details}\label{app:implementation}\label{app:actions}
\begin{table}[H]
\floatconts{tab:sizes}{\caption{Dataset partitions and severe calibration counts. N/A: no target-domain development split in the non-transfer settings; $n_H$ is the severe-case count in the safety split.}}{\small\renewcommand{\arraystretch}{0.9}\setlength{\tabcolsep}{4pt}\begin{tabular}{lrrrrrr}
\toprule
Data & Train & Classifier val. & Calibration & Target dev. & Safety $n_H$ & Test\\
\midrule
COVID-19 & 12,699 & 2,116 & 2,116 & N/A & 366 & 4,234\\
ISIC 2019 & 17,731 & 2,534 & 5,066 & N/A & 344 & 6,191\\
EyePACS$\rightarrow$APTOS & 31,612 & 3,514 & 1,464 & 732 & 65 & 1,466\\
\bottomrule
\end{tabular}}
\end{table}

\subsection*{Action Sets and Loss Matrices}
Five clinical experts designed the action categories and severe-action-error definitions: two dermatologists for ISIC, two ophthalmologists for EyePACS--APTOS, and one respiratory-medicine clinician for COVID-19. These experts specified the loss matrices for their respective tasks and confirmed the final entries shown in Table~\ref{tab:loss_matrices}.

\begin{table}[H]
\floatconts{tab:loss_matrices}{\caption{Expert-specified action loss matrices. Lower costs indicate preferred actions under each dataset's decision model.}}{%
\small\renewcommand{\arraystretch}{0.9}\setlength{\tabcolsep}{2pt}
\begin{minipage}[t]{0.43\linewidth}\centering
\textbf{(a) ISIC 2019}\par\smallskip
\begin{tabular}{lrrrr}
\toprule
Label & Routine & Referral & Biopsy & Urgent\\
\midrule
MEL & 10&4&1&0\\ NV & 0&1&3&6\\ BCC & 7&2&0&4\\ AK & 5&1&2&4\\ BKL & 0&1&3&6\\ DF & 0&1&3&6\\ VASC & 1&1&3&6\\ SCC & 9&3&1&0\\
\bottomrule
\end{tabular}
\end{minipage}\hfill
\begin{minipage}[t]{0.54\linewidth}\centering
\textbf{(b) COVID-19}\par\smallskip
\begin{tabular}{lrrrr}
\toprule
Label & No action & Antibiotics & Quarantine & Testing\\
\midrule
Normal & 0&8&8&6\\ Viral pneumonia & 10&0&7&3\\ COVID-19 & 10&7&0&2\\ Lung opacity & 9&6&6&0\\
\bottomrule
\end{tabular}
\par\smallskip
\textbf{(c) EyePACS--APTOS}\par\smallskip
\begin{tabular}{lrrrr}
\toprule
Grade & Routine & Shortened & Ophthalmology & Retina\\
\midrule
No DR & 0&1&3&6\\ Mild NPDR & 1&0&2&5\\ Moderate NPDR & 4&2&0&3\\ Severe NPDR & 8&5&2&0\\ PDR & 10&6&3&0\\
\bottomrule
\end{tabular}
\end{minipage}}
\end{table}

The escalation ordering is informed by NICE NG12 and dermatology guidance for skin lesions \citep{nice2015ng12}, and AAO guidance for retinopathy \citep{aao2019drppp}. The primary $1\%$ budget was motivated as a cross-domain reference by chest-pain risk tolerances \citep{than2013acceptable} and the ACEP policy's 30-day MACE framing \citep{tomaszewski2018acep}.

COVID-19 actions are informed by WHO management guidance \citep{who2020clinical}.

\begin{algorithm2e}[tbp]
\caption{HC-ROCP joint certification}\label{alg:protocol}
\KwIn{frozen probabilities $p(\cdot|x)$ on calibration; coverage target $1-\alpha$; harm budget $\delta$; family size $M$; groups $G$; level $\varepsilon$}
\KwOut{selected policy $\pi^*$, evidence level, subgroup statuses}
Split calibration into disjoint $\mathcal D_{\rm construct},\mathcal D_{\rm cov},\mathcal D_{\rm safety}$\;
Build candidate family $\mathcal F=\{(\beta,\gamma)\}$ on $\mathcal D_{\rm construct}$\;
$\eta\leftarrow \varepsilon/[M(1+G)]$\;
\ForEach{$\pi\in\mathcal F$}{
  $K_{\rm cov}(\pi)\leftarrow$ \#covered points of $\pi$ on $\mathcal D_{\rm cov}$\;
  $L_{\rm cov}(\pi)\leftarrow \mathrm{CP}_L(K_{\rm cov}(\pi), n_{\rm cov}, \eta)$\;
  \ForEach{group $g$}{
    $K_{H,g}(\pi)\leftarrow$ \#severe-action errors of $\pi$ on $\mathcal D_{\rm safety}$ in group $g$\;
    $U_{H,g}(\pi)\leftarrow \mathrm{CP}_U(K_{H,g}(\pi), n_{H,g}, \eta)$\;
  }
}
Define $\widehat R_{H,0}(\pi)=K_{H,0}(\pi)/n_{H,0}$ when $n_{H,0}>0$, and $\widehat R_{H,0}(\pi)=1$ when $n_{H,0}=0$ (an empty severe group is infeasible).
$\mathcal F_{\rm feas}\leftarrow\{\pi:L_{\rm cov}(\pi)\ge 1-\alpha,\ U_{H,0}(\pi)\le\delta\}$\;
$\mathcal F_{\rm exp}\leftarrow\{\pi:(K_{\rm cov}(\pi)+1)/(n_{\rm cov}+1)\ge1-\alpha,\ (K_{H,0}(\pi)+1)/(n_{H,0}+1)\le\delta\}$\;
$\mathcal F_{\rm emp}\leftarrow\{\pi:K_{\rm cov}(\pi)/n_{\rm cov}\ge1-\alpha,\ \widehat R_{H,0}(\pi)\le\delta\}$\;
\eIf{$\mathcal F_{\rm feas}\ne\emptyset$}{
  $\pi^*\leftarrow\arg\min_{\pi\in\mathcal F_{\rm feas}} \mathrm{loss}_{\mathcal D_{\rm safety}}(\pi)$\;
  level $\leftarrow$ \emph{high-confidence}\;
}{
  \eIf{$\mathcal F_{\rm exp}\ne\emptyset$}{
    $\pi^*\leftarrow\arg\min_{\pi\in\mathcal F_{\rm exp}} \mathrm{loss}_{\mathcal D_{\rm safety}}(\pi)$\;
    level $\leftarrow$ \emph{expected-risk}\;
  }{
    \eIf{$\mathcal F_{\rm emp}\ne\emptyset$}{
      $\pi^*\leftarrow\arg\min_{\pi\in\mathcal F_{\rm emp}} \mathrm{loss}_{\mathcal D_{\rm safety}}(\pi)$\;
      level $\leftarrow$ \emph{empirical}\;
    }{
      $\pi^*\leftarrow\arg\min_{\pi\in\mathcal F} \bigl[\max(0,\,1-\alpha-K_{\rm cov}(\pi)/n_{\rm cov}),\ \widehat R_{H,0}(\pi),\ \mathrm{loss}_{\mathcal D_{\rm safety}}(\pi)\bigr]$\;
      level $\leftarrow$ \emph{infeasible fallback}\;
    }
  }
}
\ForEach{subgroup $g$}{ status($g$) $\leftarrow$ \emph{certified} if $U_{H,g}(\pi^*)\le\delta_g$, else \emph{insufficient}\; }
\end{algorithm2e}

We allocate one third of calibration observations to each of the construct, coverage, and safety splits, using stratified assignment. The $\beta$ grid is obtained by coverage calibration at targets $\{0.95,0.975,0.99\}$ crossed with a log-spaced $\gamma$ grid, yielding $M=2394$ (COVID-19), $336$ (ISIC 2019), and $126$ (EyePACS--APTOS) candidates. Randomized RAPS/SAPS baselines use independent fixed random streams for calibration and test.

\paragraph{Baseline Implementations.}
We use the authors' released implementations of ROCP, RAPS, SOCOP, CRC, and LTT with the same frozen classifier probabilities. For ROCP's reference-split comparison, a single global $\beta$ is fitted by bisection to $95\%$ empirical coverage on the full calibration split. The exact, matched ROCP variant instead enumerates every nonempty label subset and is the $\gamma=0$ member of our family.

RAPS uses a randomized score with $k_{\rm reg}=1$ and $\lambda=0.01$, and SOCOP uses $\lambda=0.0625$. CRC applies its published calibration rule to the monotone action family; LTT uses Hoeffding--Bentkus $p$-values with Bonferroni correction. SAPS implements the published randomized scoring formula. Utility-directed CP uses the score-3 adapter distributed with ROCP.

\paragraph{Classifiers and Preprocessing.}
All classifiers are ConvNeXt V2-Tiny backbones \citep{woo2023convnextv2} initialized from ImageNet-22k pre-trained weights and fine-tuned per dataset (Table~\ref{tab:training}). Images are resized to $224\times224$ and normalized with ImageNet statistics; training uses random resized crops, horizontal/vertical flips, and color jitter, while validation and test use a $256$-resize followed by a center crop. Models are optimized with AdamW under a cosine learning-rate schedule, early stopping on validation balanced accuracy, and cross-entropy loss with label smoothing $0.1$. COVID-19 and ISIC use inverse-square-root class weights. The EyePACS classifier uses class-balanced sampling without loss class weights. All training and evaluation ran on a single NVIDIA RTX 3090 (24\,GB), CUDA 12.4, PyTorch 2.6, and Python 3.11.

Target-domain recalibration fits and audits HC-ROCP on the 1,464-case APTOS calibration split, with the EyePACS-trained classifier frozen.

\begin{table}[H]
\floatconts{tab:training}{\caption{Classifier fine-tuning settings.}}{\begin{tabular}{lccc}
\toprule
Setting & COVID-19 & ISIC 2019 & EyePACS$\rightarrow$APTOS\\
\midrule
Classes & 4 & 8 & 5\\
Batch size & 64 & 64 & 64\\
Max epochs & 12 & 12 & 15\\
Backbone LR & $3{\times}10^{-4}$ & $3{\times}10^{-4}$ & $1{\times}10^{-5}$\\
Head LR & $3{\times}10^{-4}$ & $3{\times}10^{-4}$ & $1{\times}10^{-4}$\\
Weight decay & 0.05 & 0.05 & 0.05\\
Label smoothing & 0.1 & 0.1 & 0.1\\
Drop path & 0.1 & 0.1 & 0.1\\
Precision & fp16 & fp16 & fp32\\
Head-only epochs & 0 & 0 & 2\\
Early-stop patience & 4 & 4 & 5\\
\bottomrule
\end{tabular}}
\end{table}

\FloatBarrier
\section{Loss-Matrix Sensitivity}\label{app:sensitivity}
Figure~\ref{fig:loss} scales only the severe-label--harmful-action loss entries by $0.5$, $1$, or $2$ and reruns the full construct/coverage/safety protocol over the same ten calibration seeds and fixed test sets as Table~\ref{tab:baselines}. The $1\times$ results reproduce Table~\ref{tab:baselines}. Each scale is a separately fixed sensitivity scenario, not selected across scales using test or certification results.

\begin{figure}[H]
\floatconts{fig:loss}{\caption{HC-ROCP matches or lowers exact ROCP's mean severe-action-error rate at all three tested cost scales. Points: mean $\pm$ sample standard deviation over ten seeds; dotted line: $1\%$ budget. Crosses mark diagnostic fallbacks with no empirically feasible candidate in any seed.}}{\includegraphics[width=0.97\linewidth]{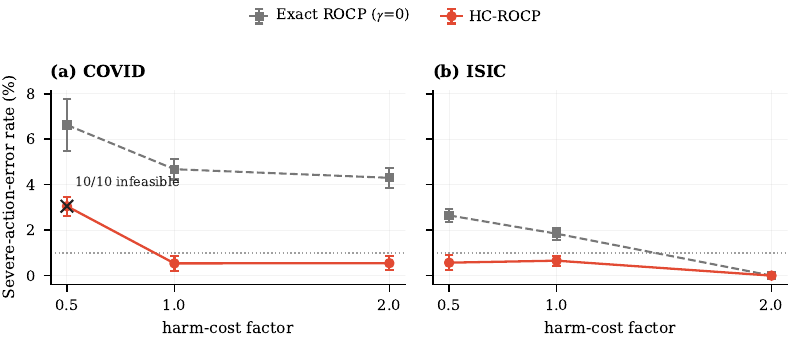}}
\end{figure}

On EyePACS--APTOS, both methods have zero test harm at all tested cost scales. COVID-19 at $1\times$/$2\times$ and ISIC at all scales retain expected-risk labels; EyePACS--APTOS retains empirical labels. No scenario yields a global or subgroup high-confidence certificate at $\delta=1\%$.

\FloatBarrier
\section{Complete Baseline Comparison and Statistical Tests}\label{app:baselines}
Table~\ref{tab:baselines} reports the full held-out test comparison averaged over ten calibration-split seeds. All methods use the same frozen probabilities and fixed test observations. Standard baselines and exact ROCP calibrate on each seed's coverage split; HC-ROCP additionally uses construct and safety splits. The comparisons therefore include the cost of data separation.

For set size, we average the number of returned labels over the fixed test set within each seed, then summarize across the same ten seeds for every set-valued method.

Table~\ref{tab:ltt} reports the action-only LTT adapter.

\begin{table}[H]
\floatconts{tab:ltt}{\caption{Official-code LTT action adapter on the same safety split as HC-ROCP. The adapter tests 101 prespecified action thresholds with the repository's Hoeffding--Bentkus $p$-value and Bonferroni correction. ``None certified'' means that no threshold passed; no policy is selected for test evaluation.}}{\begin{tabular}{lrrrrr}
\toprule
Data & $\delta$ & Certified & Threshold & Test harm & Test loss\\
\midrule
COVID-19 & 1\% & 0 & \multicolumn{3}{c}{None certified}\\
ISIC 2019 & 1\% & 0 & \multicolumn{3}{c}{None certified}\\
EyePACS$\rightarrow$APTOS & 1\% & 0 & \multicolumn{3}{c}{None certified}\\
COVID-19 & 5\% & 16 & 0.15 & 0.68\% & 0.954\\
ISIC 2019 & 5\% & 7 & 0.06 & 1.01\% & 1.193\\
EyePACS$\rightarrow$APTOS & 5\% & 0 & \multicolumn{3}{c}{None certified}\\
\bottomrule
\end{tabular}}
\end{table}

\paragraph{Statistical Tests.}
We use ten stratified calibration partitions (seeds 20260815--20260824). Two-sided paired $t$-tests compare seed-level severe-action-error rate and loss. Holm correction is applied separately within each dataset/metric over the 13 eligible baselines. Identical paired values receive $p=1$. Tests quantify calibration-partition variation conditional on these fixed data and classifier, not new training runs or independent test cohorts. The ISIC release-solver harm value is $0.47\%\pm0.00\%$ because all ten coverage-split recalibrations select the same test action pattern.
\paragraph{Practical Significance.} Before these comparisons, we specified minimum effect sizes of 1 percentage point for harm (the primary $\delta$) and 0.1 for loss.
For ISIC's release solver versus HC-ROCP, the Holm-adjusted values are $p=0.0679$ for harm and $p=0.0171$ for loss. The mean loss advantage is $0.0546$, below the $0.1$ minimum-effect threshold.
\paragraph{Reporting.} A baseline receives $\dagger$ if HC-ROCP is better and $\ddagger$ if the baseline is better, requiring both adjusted $p<0.05$ and the relevant minimum effect size. Coverage is reported descriptively. Absence of a mark does not establish equivalence. Evidence labels and certificate counts are not tested. Tables~\ref{tab:matched} and~\ref{tab:baselines} use the same tests.

\begin{table}[p]
\floatconts{tab:baselines}{\caption{Ten-seed held-out comparison (mean $\pm$ sample standard deviation). Harm: severe-action-error rate (\%); loss: modeled decision loss; set size: mean label count. $\dagger$/$\ddagger$: baseline harm/loss significantly worse/better than HC-ROCP (paired Holm $p<0.05$ and minimum effect size; Appendix~\ref{app:baselines}). N/A: not applicable to action-only methods, which return no prediction set.}}{\small\renewcommand{\arraystretch}{0.89}\begin{tabular}{lrrrr}
\toprule
\multicolumn{5}{l}{\textbf{(a) COVID-19}}\\
Method & Coverage & Harm (\%) & Loss & Set size\\
\midrule
ROCP (release solver) & $0.952\pm0.004$ & $4.45\pm0.13^{\dagger}$ & $0.391\pm0.005^{\ddagger}$ & $1.020\pm0.023$\\
ROCP (exact, matched) & $0.952\pm0.006$ & $4.68\pm0.45^{\dagger}$ & $0.391\pm0.010^{\ddagger}$ & $1.019\pm0.029$\\
RAC & $0.953\pm0.004$ & $5.10\pm0.40^{\dagger}$ & $0.389\pm0.006^{\ddagger}$ & $1.020\pm0.030$\\
LAS & $0.951\pm0.006$ & $4.86\pm0.31^{\dagger}$ & $0.388\pm0.009^{\ddagger}$ & $1.005\pm0.014$\\
APS & $0.953\pm0.004$ & $5.01\pm0.42^{\dagger}$ & $0.387\pm0.008^{\ddagger}$ & $1.008\pm0.010$\\
RAPS (rand., official) & $0.943\pm0.008$ & $2.69\pm0.45^{\dagger}$ & $0.749\pm0.141^{\ddagger}$ & $1.087\pm0.040$\\
SAPS (rand., formula) & $0.958\pm0.008$ & $0.56\pm0.46$ & $1.527\pm0.607$ & $1.514\pm0.161$\\
Mondrian APS & $0.963\pm0.004$ & $3.03\pm0.69^{\dagger}$ & $0.445\pm0.056^{\ddagger}$ & $1.044\pm0.019$\\
SOCOP (official) & $0.952\pm0.006$ & $4.86\pm0.33^{\dagger}$ & $0.385\pm0.008^{\ddagger}$ & $1.018\pm0.031$\\
Utility-directed CP (score adapter) & $0.953\pm0.007$ & $0.00\pm0.00$ & $3.422\pm0.000^{\dagger}$ & $3.297\pm0.096$\\
Bayes action & N/A & $4.46\pm0.00^{\dagger}$ & $0.391\pm0.000^{\ddagger}$ & N/A\\
Always-safe action & N/A & $0.00\pm0.00$ & $2.935\pm0.000^{\dagger}$ & N/A\\
CRC action (official) & N/A & $0.30\pm0.26$ & $1.360\pm0.280$ & N/A\\
HC-ROCP & $0.989\pm0.002$ & $0.54\pm0.32$ & $1.181\pm0.337$ & $1.732\pm0.185$\\
\midrule
\multicolumn{5}{l}{\textbf{(b) ISIC 2019}}\\
Method & Coverage & Harm (\%) & Loss & Set size\\
\midrule
ROCP (release solver) & $0.942\pm0.004$ & $0.47\pm0.00$ & $1.475\pm0.018$ & $5.755\pm0.102$\\
ROCP (exact, matched) & $0.943\pm0.005$ & $1.84\pm0.27^{\dagger}$ & $1.478\pm0.016$ & $5.739\pm0.131$\\
RAC & $0.943\pm0.005$ & $3.44\pm0.31^{\dagger}$ & $1.505\pm0.011$ & $5.725\pm0.118$\\
LAS & $0.942\pm0.004$ & $0.57\pm0.04$ & $1.906\pm0.004^{\dagger}$ & $4.790\pm0.198$\\
APS & $0.943\pm0.003$ & $0.34\pm0.06$ & $1.912\pm0.002^{\dagger}$ & $4.622\pm0.147$\\
RAPS (rand., official) & $0.939\pm0.004$ & $0.05\pm0.07$ & $1.905\pm0.006^{\dagger}$ & $4.385\pm0.137$\\
SAPS (rand., formula) & $0.949\pm0.006$ & $1.01\pm0.28$ & $1.884\pm0.007^{\dagger}$ & $6.203\pm0.167$\\
Mondrian APS & $0.941\pm0.005$ & $0.00\pm0.00$ & $1.904\pm0.012^{\dagger}$ & $5.707\pm0.162$\\
SOCOP (official) & $0.948\pm0.005$ & $1.66\pm0.24^{\dagger}$ & $1.836\pm0.005^{\dagger}$ & $6.202\pm0.147$\\
Utility-directed CP (score adapter) & $0.941\pm0.003$ & $0.06\pm0.02$ & $1.982\pm0.001^{\dagger}$ & $5.788\pm0.155$\\
Bayes action & N/A & $3.22\pm0.00^{\dagger}$ & $1.137\pm0.000^{\ddagger}$ & N/A\\
Always-safe action & N/A & $0.00\pm0.00$ & $1.249\pm0.000^{\ddagger}$ & N/A\\
CRC action (official) & N/A & $0.66\pm0.73$ & $1.216\pm0.036^{\ddagger}$ & N/A\\
HC-ROCP & $0.956\pm0.012$ & $0.66\pm0.22$ & $1.530\pm0.059$ & $6.172\pm0.404$\\
\midrule
\multicolumn{5}{l}{\textbf{(c) EyePACS$\rightarrow$APTOS}}\\
Method & Coverage & Harm (\%) & Loss & Set size\\
\midrule
ROCP (release solver) & $0.949\pm0.007$ & $0.00\pm0.00$ & $1.381\pm0.067^{\ddagger}$ & $3.836\pm0.137$\\
ROCP (exact, matched) & $0.950\pm0.005$ & $0.00\pm0.00$ & $1.514\pm0.051^{\ddagger}$ & $3.994\pm0.103$\\
RAC & $0.951\pm0.005$ & $0.00\pm0.00$ & $1.390\pm0.057^{\ddagger}$ & $3.926\pm0.097$\\
LAS & $0.936\pm0.005$ & $0.00\pm0.00$ & $1.315\pm0.036^{\ddagger}$ & $3.455\pm0.069$\\
APS & $0.938\pm0.006$ & $0.00\pm0.00$ & $1.356\pm0.074^{\ddagger}$ & $3.431\pm0.096$\\
RAPS (rand., official) & $0.937\pm0.011$ & $0.00\pm0.00$ & $1.302\pm0.117^{\ddagger}$ & $3.330\pm0.142$\\
SAPS (rand., formula) & $0.953\pm0.006$ & $0.00\pm0.00$ & $1.995\pm0.004^{\dagger}$ & $4.597\pm0.035$\\
Mondrian APS & $0.939\pm0.007$ & $0.00\pm0.00$ & $1.834\pm0.173^{\dagger}$ & $3.663\pm0.220$\\
SOCOP (official) & $0.955\pm0.005$ & $0.00\pm0.00$ & $1.983\pm0.005^{\dagger}$ & $4.534\pm0.063$\\
Utility-directed CP (score adapter) & $0.935\pm0.009$ & $0.00\pm0.00$ & $2.027\pm0.000^{\dagger}$ & $4.404\pm0.085$\\
Bayes action & N/A & $0.00\pm0.00$ & $1.323\pm0.000^{\ddagger}$ & N/A\\
Always-safe action & N/A & $0.00\pm0.00$ & $1.415\pm0.000^{\ddagger}$ & N/A\\
CRC action (official) & N/A & $0.00\pm0.00$ & $1.415\pm0.000^{\ddagger}$ & N/A\\
HC-ROCP & $0.967\pm0.006$ & $0.00\pm0.00$ & $1.705\pm0.081$ & $4.351\pm0.139$\\
\bottomrule
\end{tabular}}
\end{table}

\paragraph{Severe-Subgroup Results.}
Figure~\ref{fig:matrix} reports severe-subgroup status at each budget.

\begin{figure}[H]
\floatconts{fig:matrix}{\caption{Only MEL ($n=302$) certifies, at $\delta=5\%$, across the tested subgroup--budget pairs. Each cell reports the Clopper--Pearson upper bound $U_g$ and is colored by $\delta-U_g$ (rust: insufficient; green: certified). Lollipops show subgroup safety-calibration counts.}}{\includegraphics[width=\linewidth]{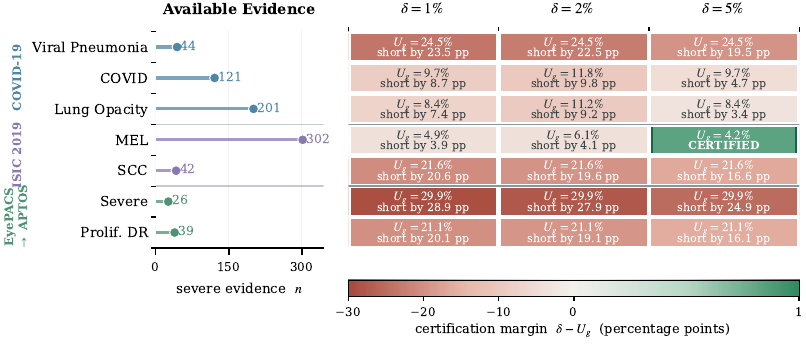}}
\end{figure}

\FloatBarrier
\section{Protocol Diagnostics and Parameter Sensitivity}\label{app:ablation}
Table~\ref{tab:ablation} compares protocol variants on frozen COVID-19 probabilities. The $\gamma=0$ exact-ROCP row uses 1,410 calibration cases to fit $\beta$, whereas the full protocol constructs its candidate family on 705 cases and uses separate coverage and safety splits. The change from $4.51\%$ to $0.50\%$ harm is therefore a protocol-level comparison, not an isolated causal effect of adding $\gamma$.

The fixed-$\beta$ diagnostics use the full protocol's construct and safety observations, but change the selection rule. Fitting $\beta$ at $\gamma=0$ and choosing the first $\gamma$ meeting the safety-count criterion yields $0.786$ test coverage. Taking the largest construct-calibrated $\beta$ across the grid yields loss $1.405$.

Reusing calibration observations changes the selected operating point but produces no high-confidence certificate in this experiment. This empirical outcome does not validate the reuse: independence between construction and certification is a condition of Theorem~\ref{thm:joint}. On ISIC, dropping the subgroup multiplicity factor tightens the SCC upper bound from $21.6\%$ to $18.9\%$, still above $5\%$.

\begin{table}[H]
\floatconts{tab:ablation}{\caption{Protocol diagnostics on frozen COVID-19 probabilities. Level reports descriptive evidence for the split, multiplicity, and grid variants, and diagnostic outcomes for the fixed-$\beta$ and $\gamma=0$ rows.}}{\begin{tabular}{lrrrr}
\toprule
Protocol variant & Cov. & Loss & Harm (\%) & Level\\
\midrule
Full (ours) & 0.990 & 1.080 & 0.50 & expected-risk\\
Na\"ive single-split & 0.989 & 0.973 & 0.68 & expected-risk\\
Two-split (construct$=$coverage) & 0.990 & 1.080 & 0.50 & expected-risk\\
Two-split (coverage$=$safety) & 0.990 & 1.042 & 0.59 & expected-risk\\
No subgroup multiplicity & 0.990 & 1.080 & 0.50 & expected-risk\\
Fixed-$\beta$ (sweep $\gamma$) & 0.786 & 1.285 & 0.32 & coverage collapse\\
Fixed-$\beta$ (max over grid) & 0.959 & 1.405 & 0.23 & over-conservative\\
$\gamma{=}0$ (exact ROCP) & 0.947 & 0.394 & 4.51 & no harm control\\
Sparse grid ($M{=}40$) & 0.949 & 2.950 & 0.00 & expected-risk\\
Dense grid ($M{=}135$) & 0.991 & 1.350 & 0.27 & expected-risk\\
\bottomrule
\end{tabular}}
\end{table}

Table~\ref{tab:shortlist} isolates candidate screening by starting from the full $M=2394$ family. Using only the construct split, we rank candidates by expected-coverage shortfall, expected-harm excess, and utility, then freeze the top $L$ before testing. We evaluate prespecified $L\in\{1,5,10,20,50,100,250,500,1000\}$ separately, with correction $L(1+G)$; the table shows representative sizes. Smaller families tighten the bounds, but no shortlist certifies at $1\%$. Even one candidate requires more severe cases than are available under zero errors. Aggressive screening can discard the policy recovered by larger shortlists, exposing a trade-off between candidate quality and multiplicity.

\begin{table}[H]
\floatconts{tab:shortlist}{\caption{Construct-only shortlist ablation on COVID-19 at $\delta=1\%$. $n_0$: minimum severe-case count for the zero-error global harm gate at each row's multiplicity. $U_H$: selected policy's global harm Clopper--Pearson upper bound on the disjoint safety split.}}{\begin{tabular}{rrrrrrr}
\toprule
$L$ & $\eta$ & $n_0$ & Safety $k/n$ & $U_H$ & Evidence & Test loss / harm\\
\midrule
1    & $1.00\mathrm{e}{-2}$ & 459  & $4/366$ & 3.14\% & infeasible & 0.972 / 0.68\%\\
10   & $1.00\mathrm{e}{-3}$ & 688  & $4/366$ & 3.98\% & infeasible & 0.972 / 0.68\%\\
50   & $2.00\mathrm{e}{-4}$ & 848  & $4/366$ & 4.54\% & infeasible & 0.972 / 0.68\%\\
250  & $4.00\mathrm{e}{-5}$ & 1008 & $2/366$ & 4.02\% & expected-risk & 1.080 / 0.50\%\\
1000 & $1.00\mathrm{e}{-5}$ & 1146 & $2/366$ & 4.43\% & expected-risk & 1.080 / 0.50\%\\
2394 & $4.18\mathrm{e}{-6}$ & 1233 & $2/366$ & 4.69\% & expected-risk & 1.080 / 0.50\%\\
\bottomrule
\end{tabular}}
\end{table}

Table~\ref{tab:alpha} reports the coverage-target sensitivity of the recalibrated candidate family. Each row recalibrates its own candidate family, so the candidate sets need not be nested across targets. The $0.95$ and $0.99$ targets select the same $(\beta,\gamma)$ pair. At the looser $0.90$ target, selection instead returns $\beta=6.17$ and $\gamma=31.6$, with loss $1.190$ rather than $1.080$. Thus loss is non-monotone across these separately constructed families.

\begin{table}[H]
\floatconts{tab:alpha}{\caption{Coverage-target ($\alpha$) sensitivity on COVID-19 with separately recalibrated candidate families. All selected policies have expected-risk status and $0.50\%$ test harm.}}{\begin{tabular}{lrrrrrr}
\toprule
$\alpha$ & Target & Cov. & Loss & Harm (\%) & $\beta$ & $\gamma$\\
\midrule
0.10 & 0.90 & 0.906 & 1.190 & 0.50 & 6.17 & 31.6\\
0.05 & 0.95 & 0.990 & 1.080 & 0.50 & 35.7 & 1.78\\
0.01 & 0.99 & 0.990 & 1.080 & 0.50 & 35.7 & 1.78\\
\bottomrule
\end{tabular}}
\end{table}

\paragraph{Count Diagnostic.} The real-data series in Figure~\ref{fig:power}(a) uses the already selected COVID-19 policy and subsamples its severe safety cases, then applies an unadjusted single-test harm bound. The diagnostic uses 20 subsamples each at $n_H=50,100,200$ and the single full safety sample at $n_H=366$; all gate-pass rates are zero. Because selection used the same safety data, this is an exploratory count diagnostic, not a fresh certification test or an independent estimate of the complete protocol's power.

\paragraph{Nested Bootstrap.}
For each of 500 repetitions at $\delta=1\%$, we resample the full calibration set's frozen probability--label pairs with replacement within label strata and independently bootstrap the test pairs without stratification (base seed 20260821).
We then redraw the three stratified calibration splits, reconstruct the candidate family, repeat all coverage and harm tests, select a policy, and evaluate it on the resampled test set.
COVID-19 and ISIC retain expected-risk labels in $500/500$ repetitions each, and EyePACS--APTOS retains its empirical label in $500/500$.

\section{Simplified False-Certification Simulation}\label{app:protocolstudy}
We report a simplified simulation of global false-certification rates under different multiplicity and search assumptions (Figure~\ref{fig:protocolstudy}). With full correction, the observed rate ranges from $0\%$ to $0.25\%$, below the $5\%$ nominal level. We simulate $M=100$ candidates with true harm $1.1\delta$ at $\delta=1\%$, using 400 Monte Carlo repetitions per sample size for each of the three settings. The plotted event is at least one false global-harm claim among the candidates.

Full correction uses $\eta=\varepsilon/[M(1+G)]$ with $\varepsilon=0.05$ and $G=4$. Reduced correction uses $\eta=\varepsilon/M$. The expanded-search curve tests $25M$ candidates at the full-correction level computed for $M$. Each setting and sample size uses a separate random stream derived from seed 20261108. These settings isolate correction and search-size assumptions in binomial harm tests; the coverage gate and data-dependent policy construction are outside this simulation.

For the complete accounting of $M(1+G)$ tests, replacing the per-test allowance by $\varepsilon/M$ loosens the union-bound ceiling to $\varepsilon(1+G)=25\%$. This is a bound on error probability, rather than a predicted or attained rate.

\begin{figure}[H]
\floatconts{fig:protocolstudy}{\caption{Global false-certification rate under three correction and search settings. Vertical markers show reference severe-case counts: SCC (42), EyePACS--APTOS (65), and COVID-19 (366).}}{\includegraphics[width=0.92\linewidth]{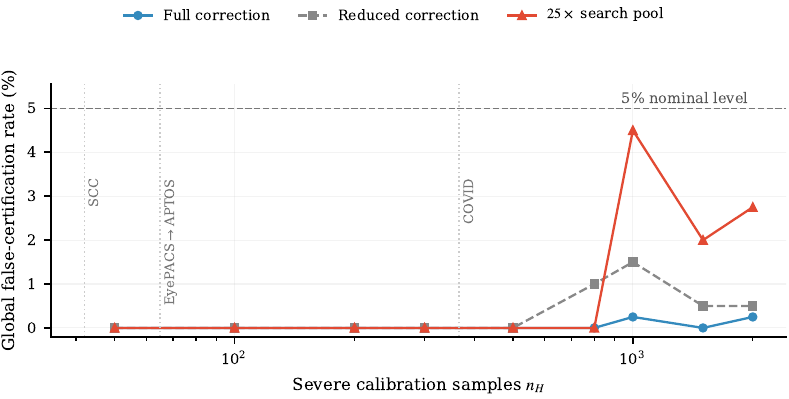}}
\end{figure}

\FloatBarrier
\noindent\begin{minipage}{\linewidth}
\section{Referral Actions and the Action-Space Gap}\label{app:refer}
Adding referral tests how the action space affects the COVID-19 trade-off. The original actions are no action, antibiotics, quarantine, and additional testing. We append a ``refer to human'' action with fixed modeled cost $c$, classify referral as non-harmful, and rerun the joint protocol on the reference split. Across $c\in[0.3,1.0]$, test loss falls from $1.080$ to $0.28$--$0.40$, while the severe-action-error rate stays at or below $0.50\%$ (Table~\ref{tab:refer}). Referral volume varies from $22\%$ to $100\%$: at $c=0.3$, the policy refers every case and has loss $0.300$. The harm rate is non-monotone in $c$ because selection minimizes calibration loss among expected-feasible candidates rather than minimizing test harm.

\begin{table}[H]
\floatconts{tab:refer}{\caption{COVID-19 referral-cost sweep. N/A: the original action set has no referral option.}}{\begin{tabular}{lrrrr}
\toprule
Action set & Cov. & Loss & Harm (\%) & Referred\\
\midrule
Original (4 actions) & 0.990 & 1.080 & 0.50 & N/A\\
$+$Refer ($c{=}0.3$) & 1.000 & 0.300 & 0.00 & 100\%\\
$+$Refer ($c{=}0.4$) & 0.991 & 0.342 & 0.00 & 74\%\\
$+$Refer ($c{=}0.5$) & 0.993 & 0.326 & 0.00 & 56\%\\
$+$Refer ($c{=}0.6$) & 0.977 & 0.288 & 0.18 & 25\%\\
$+$Refer ($c{=}0.7$) & 0.979 & 0.284 & 0.36 & 22\%\\
$+$Refer ($c{=}0.8$) & 0.982 & 0.355 & 0.14 & 30\%\\
$+$Refer ($c{=}1.0$) & 0.983 & 0.399 & 0.14 & 29\%\\
\bottomrule
\end{tabular}}
\end{table}

\end{minipage}
\end{document}